\documentclass[lettersize,journal]{IEEEtran}
\usepackage{amsmath,amsfonts}
\usepackage{algorithmic}
\usepackage{algorithm}
\usepackage{array}
\usepackage[caption=false,font=normalsize,labelfont=sf,textfont=sf]{subfig}
\usepackage{textcomp}
\usepackage{stfloats}
\usepackage{url}
\usepackage{verbatim}
\usepackage{graphicx}
\usepackage{cite}

\usepackage{booktabs}

\begin{document}

\title{PAPooling: Graph-based Position Adaptive Aggregation of Local Geometry in Point Clouds}

\author{Jie Wang, Jianan Li, Lihe Ding, Ying Wang, Tingfa Xu

\thanks{Jie Wang, Jianan Li, Lihe Ding, Ying Wang, Tingfa Xu are with School of Optics and Photonics, Image Engineering \& Video Technology Lab, Beijing Institute of Technology, Beijing 100081, China and the Key Laboratory of Photoelectronic Imaging Technology and System, Ministry of Education of China, Beijing, 100081, China. e-mail: jwang991020@gmail.com (Jie Wang), lijianan@bit.edu.cn (Jianan Li), dean.dinglihe@outlook.com (Lihe Ding), wangying7275@gmail.com (Ying Wang), ciom\_xtf1@bit.edu.cn (Tingfa Xu).}
\thanks{Tingfa Xu is with Big Data and Artificial Intelligence Laboratory, Beijing Institute of Technology Chongqing Innovation Center (BITCQIC), Chongqing, 401135, China.}
\thanks{Jianan Li and Tingfa Xu are the corresponding authors.}

}

\markboth{Journal of \LaTeX\ Class Files,~Vol.~14, No.~8, August~2021}%
{Shell \MakeLowercase{\textit{et al.}}: A Sample Article Using IEEEtran.cls for IEEE Journals}


\maketitle

\begin{abstract}
Fine-grained geometry, captured by aggregation of point features in local regions, is crucial for object recognition and scene understanding in point clouds. Nevertheless, existing preeminent point cloud backbones usually incorporate max/average pooling for local feature aggregation, which largely ignores points' positional distribution, leading to inadequate assembling of fine-grained structures. 
To mitigate this bottleneck, we present an efficient alternative to max pooling, Position Adaptive Pooling (PAPooling), that explicitly models spatial relations among local points using a novel graph representation, and aggregates features in a position adaptive manner, enabling position-sensitive representation of aggregated features.  
Specifically, PAPooling consists of two key steps, \textit{Graph Construction} and \textit{Feature Aggregation}, respectively in charge of constructing a graph with edges linking the center point with every neighboring point in a local region to map their relative positional information to channel-wise attentive weights, and adaptively aggregating local point features based on the generated weights through Graph Convolution Network (GCN). 
PAPooling is simple yet effective, and flexible enough to be ready to use for different popular backbones like PointNet++ and DGCNN, as a plug-and-play operator. 
Extensive experiments on various tasks ranging from 3D shape classification, part segmentation to scene segmentation 
well demonstrate that PAPooling can significantly improve predictive accuracy, while with minimal extra computational overhead.
Code will be released.
\end{abstract}

\begin{IEEEkeywords}
Point Cloud, point-based network, feature aggregation operation, geometric structures, graph convolution network, plug-and-play operator  
\end{IEEEkeywords}

\section{Introduction}
\IEEEPARstart{W}{ith} the advance of 3D acquisition technology, sensors like 3D scanner and LiDAR have become increasingly common and accessible.
3D point clouds obtained by these sensors provide rich geometric, shape and scale information, and play a crucial role in the field of robots~\cite{referpaper2,referpaper21}, autonomous driving~\cite{referpaper9}, virtual and augmented reality~\cite{referpaper1,referpaper19} increasingly. 
However, in view of its inherent sparsity, irregularity and disorder characteristics, how to extract fine-grained geometry underlying point clouds that is crucial to accurate object recognition and precise scene understanding still remains a key but challenging problem.

Recently, deep learning approaches have made remarkable progress in processing point cloud data~\cite{qi2017pointnet,qi2017pointnet++,thomas2019kpconv,lin2020fpconv,DensePoint,wang2019dynamicDGCNN}, among which PointNet~\cite{qi2017pointnet}, a pioneering work of point-based model, directly works on irregular and disordered points without projection or voxelization and thus achieves ideal accuracy-speed trade-off. 
Since then, many advanced point-based models have emerged, e.g., PointNet++~\cite{qi2017pointnet++} and DGCNN~\cite{wang2019dynamicDGCNN}, which become the mainstream frameworks in handling point clouds.

\begin{figure}[!t]
	\centering
	\includegraphics[width=1.0\columnwidth]{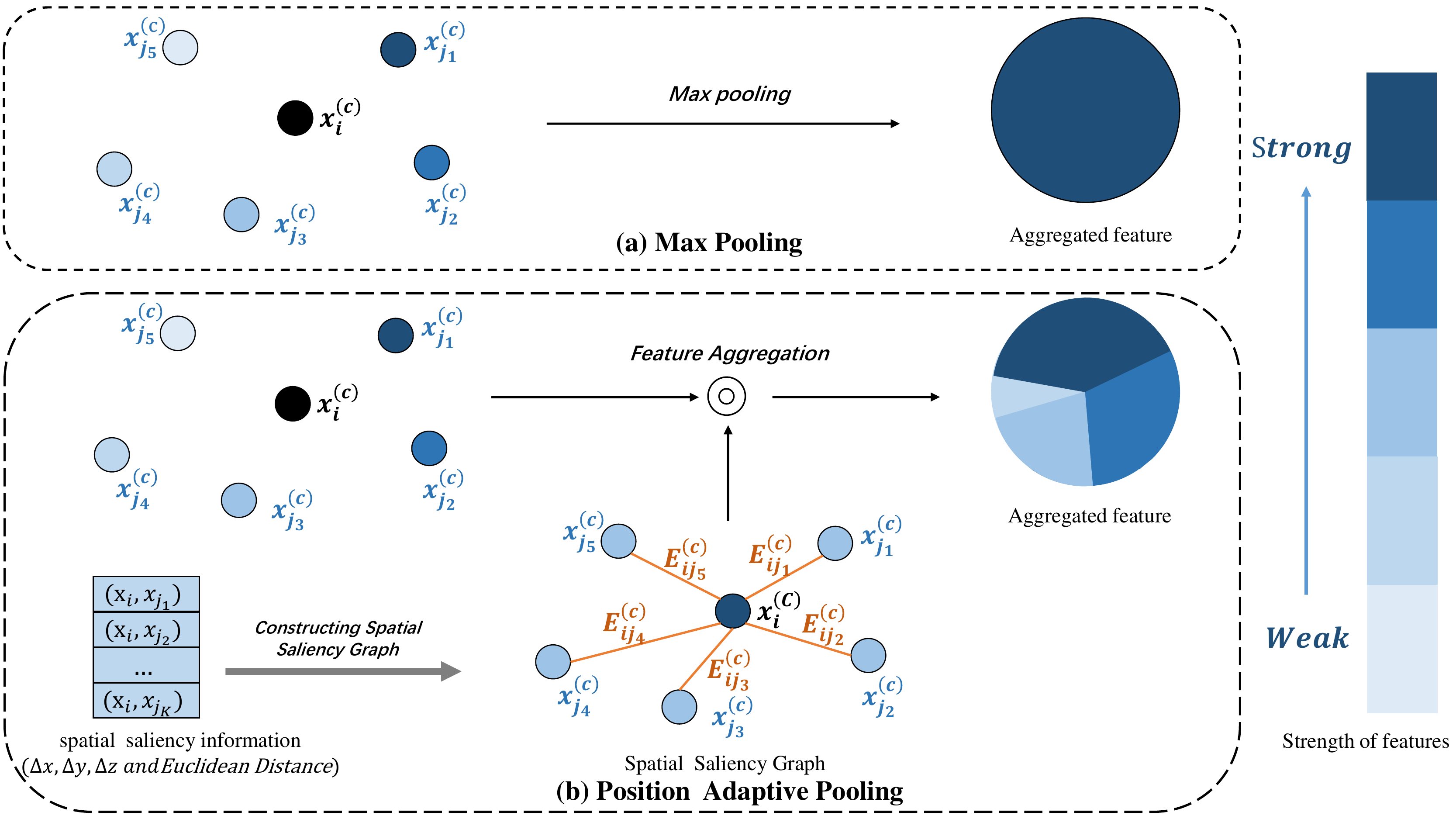} 
	\caption{Comparison of (a) max pooling and (b) our PAPooling. The left blue circles represent points in a local region (darker color corresponds to stronger feature response) and the right circle represents aggregated feature. Max pooling simply ignores the spatial distribution of local points and  retains only the features with the strongest response from a single point. In contrast, PAPooling maps the relative position information of every neighboring point with respect to the center to attentive weights by a graph representation, and aggregates features from all neighboring points in a position adaptive manner.}
	\label{comparison of max pooling and PAPooling}
\end{figure}

Significantly, feature aggregation operation, which integrates embedded features of points in a local region into an abstracted local feature descriptor, serves as a key operation in learning local geometry for point-based models.
While conventional aggregation operations like max or average pooling ignore positional distribution of points, leading to inadequate awareness of fine-grained geometry. As depicted in Fig.\ref{comparison of max pooling and PAPooling} (a), the aggregated local feature from max pooling (the right circle) is completely occupied by the single point feature with the strongest response (the left darkest circle), while the relative positional information among points is simply lacking.

To mitigate this issue, inspired by the fact that a graph is superior in modeling relationships amongst nodes, we consider the feasibility of transferring points in a local region into the form of graph, with edges encoding the spatial relations amongst points. As such, local geometric relationship can be well retained to facilitate feature aggregation in a position-sensitive manner. Following the above motivation, we originally present innovative Position Adaptive Aggregation, PAPooling, a unique plug-and-play pooling operation based on graph representation. 


We decompose the process of PAPooling into two steps: i) \textit{Graph Construction} that transforms points in a local region into a graph by linking the center and its neighboring points and by taking their relative positional information as edge features, which are further embedded into channel-wise attentive weights through Graph Convolution Network (GCN)~\cite{GCNkipf2016semi}; 2) \textit{Feature Aggregation} that computes local feature descriptor as a weighted sum of the point features at all nodes using the generated per-channel weights.
Hence, PAPooling mitigates the problem that conventional feature aggregation easily loses fine-grained geometric structures in an ingenious way.
As demonstrated in Fig.\ref{comparison of max pooling and PAPooling} (b), point features from different positions (the left circles) contribute to the aggregated feature (the right circle) with different degrees depending to their relative positions to the center.
By incorporating PAPooling into existing preeminent point-based models, fine-grained local geometry can be well preserved to learn densely contextual representation for point cloud processing.

We verify the effectiveness of PAPooling based on various popular point-based models on three challenging tasks, i.e., 3D shape classification (ModelNet40~\cite{referpaper27modelnet40}), part segmentation (ShapeNet~\cite{referpaper28_shapenet}), and semantic segmentation (S3DIS~\cite{referpaper29_s3dis}). To investigate the effect of pooling operation, we simply replace max pooling in backbones like PointNet++~\cite{qi2017pointnet++} and DGCNN~\cite{wang2019dynamicDGCNN} with PAPooling, without changing other network configurations. Experimental results show that our method improves the performance of PointNet++ by $2.0\%$ on ModelNet40, and improves the baseline of DGCNN by $1.5\%$ on ShapeNet and $4.2\%$ on S3DIS, while introducing minimal computational cost. 


To sum up, the contributions of this work are as follows: 
\begin{itemize}
\item We propose a Point-to-Graph mechanism to preserve fine-grained local geometry by taking advantage of the best relation modeling characteristics of graph representation, paving a new way to aggregating point features in a position adaptive fashion.

\item We present a plug-and-play feature aggregation operation, PAPooling, which is simple yet effective, and flexible enough to be incorporated into popular point-based models to lift their representational learning capability for point cloud processing.

\item PAPooling achieves competing performance on various challenging benchmark datasets.
\end{itemize}

\section{Related Work}
\subsection{View-based and voxel-based methods}
To replicate the success of neural networks on 2D-CNN, view-based and voxel-based methods are proposed to process raw point cloud~\cite{MVCNN}.
View-based methods convert 3D point cloud data into a set of 2D images from various angles, then use deep learning to deal with these projected multi-view pictures~\cite{guo2016multi,gvcnn,deepshape, qi2016volumetric}. They often fine tune a pre-trained image-based architecture for accurate recognition. However, 3D objects are self-occluded, which will cause certain information loss. 
To retain the information of raw point cloud as much as possible, voxel-based methods quantize 3D space and map points into regular voxels, then making point cloud convoluted with three-dimensional meshes~\cite{referpaper10voxnet,riegler2017octnet,referpaper12_splatnet}, e.g., Voxnet~\cite{referpaper10voxnet}. 
However, volumetric representation will introduce important geometric information loss due to quantization of different levels, and voxelization usually brings extra storage and calculation costs.
Advanced adaptive methods such as k-d tress~\cite{klokov2017escape} and octrees~\cite{riegler2017octnet,tatarchenko2017octree} reduce the memory and computational cost of 3D CNN, while still suffering from the trade-off between the quantization and the computational efficiency.
Yet, both the two streams require to preprocess the input point cloud and may damage the geometric information.

\subsection{Point-based methods}
Due to its irregular format and permutation invariance problem, how to process unstructured representation is one of the greatest challenges to point cloud analysis. PointNet~\cite{qi2017pointnet} is the pioneering point-based method working on irregular and disordered points, via point-wise $1\times1$ Multi-Layer Perceptrons(MLPs)~\cite{hornik1991approximationMLP} followed by a global max pooling layer to aggregate information. 

Inspired by PointNet, follow-up works have explored the point-based methods to various directions.
PointNet++~\cite{qi2017pointnet++} proposes a hierarchical structure by applying PointNet to local areas, showing the properties of local areas improve deep learning on 3D point clouds indeed. A-CNN~\cite{A-CNNkomarichev2019cnn} presents a novel annular convolution operator which can well capture the local neighborhood geometry of each point by specifying the ring-shaped structures and directions in the computation. Gird-GCN~\cite{grid-gcnxu2020grid} presents an innovative data structuring strategy Coverage-Aware Grid Query (CAGQ), improving spatial coverage while reducing the theoretical time complexity. PointConv~\cite{PointConv} reweights the convolution kernel by the density of points, able to fully approximating the 3D continuous convolution on any set of 3D points. PointCNN~\cite{PointCNN} leverages spatially-local correlation in data represented densely in grids, learns an X-transformation from the input points and then applies element-wise product and sum operations of the typical convolution operator on the X-transformed features. By means of this generalization of typical CNNs, PointCNN learns features from point clouds well. DGCNN~\cite{wang2019dynamicDGCNN} is an excellent architecture that exploits local geometric structures by constructing a local neighborhood graph, inspiring a new stream for processing point cloud. PAConv~\cite{xu2021paconv}, a position adaptive convolution operator with dynamic kernel assembling for point cloud processing, extracts the spatial information of point cloud.



\subsection{Feature aggregation in point cloud.} 
Feature aggregation is an important component of architectures designed for point cloud, playing a key role for point cloud similar to convolution kernel for image pixels. How to develop effective feature aggregation operators is a key task of point representation analysis. Recently, much effort has been made for effective local feature aggregation.  Point2Sequence~\cite{liu2019point2sequence} and A-SCN~\cite{xie2018attentional_A-SCN} introduce point attention mechanism~\cite{wang2018non-local} in the use of recurrent neural network (RNN~\cite{zaremba2014recurrent_rnn}) based encoder-decoder structure, highlighting the importance of different area scales. DGCNN~\cite{wang2019dynamicDGCNN} proposes EdgeConv module to exploiting relations between key points and their feature-space neighbors, obtaining much more fine-grained features after aggregation layers. LPD-Net~\cite{liu2019lpd-net} extends DGCNN on both spatial neighbors and feature neighbors for aggregation. RS-CNN~\cite{liu2019relation_rscnn} and KPConv~\cite{thomas2019kpconv} take consideration of spatial distribution of data, and originally propose a spatially deformable point convolution that enables model better capability of aggregating features.

Most exploratory studies for feature aggregation in point cloud tend to construct modules capable of capturing more abundant information in the point-wise transformation layer, and features aggregated become much more fine-grained of itself after normal max/average pooling. Recently, there exists some works attaching to designing novel feature aggregation operator to directly substitute max/average pooling.  RandLA-Net~\cite{hu2020randla-net} originally introduces attention mechanism~\cite{wang2018non-local} to pooling layer and proposes Attentive Pooling, generating informative feature vectors instead of using max pooling to hard integrate the neighbouring features. PosPool~\cite{liu2020closer_pospool} simply combines the relative position and point feature by element-wise multiplication without learnable weights, performs well on benchmarks.  





Attentive Pooling~\cite{hu2020randla-net} is a suggestive try for aggregation operator, while using shared weights for weighted summation, unable to realize the full potential of attention mechanism~\cite{wang2018non-local}. PosPool~\cite{liu2020closer_pospool} is a simple yet effective method, while lacks the sufficient leverage of position. By contrast, our work designs modules corresponding to shortage above and performs better in learning contextual shape-aware representation in feature aggregation process.

\begin{figure*}[ht]
\centering
\includegraphics[width=0.98\textwidth]{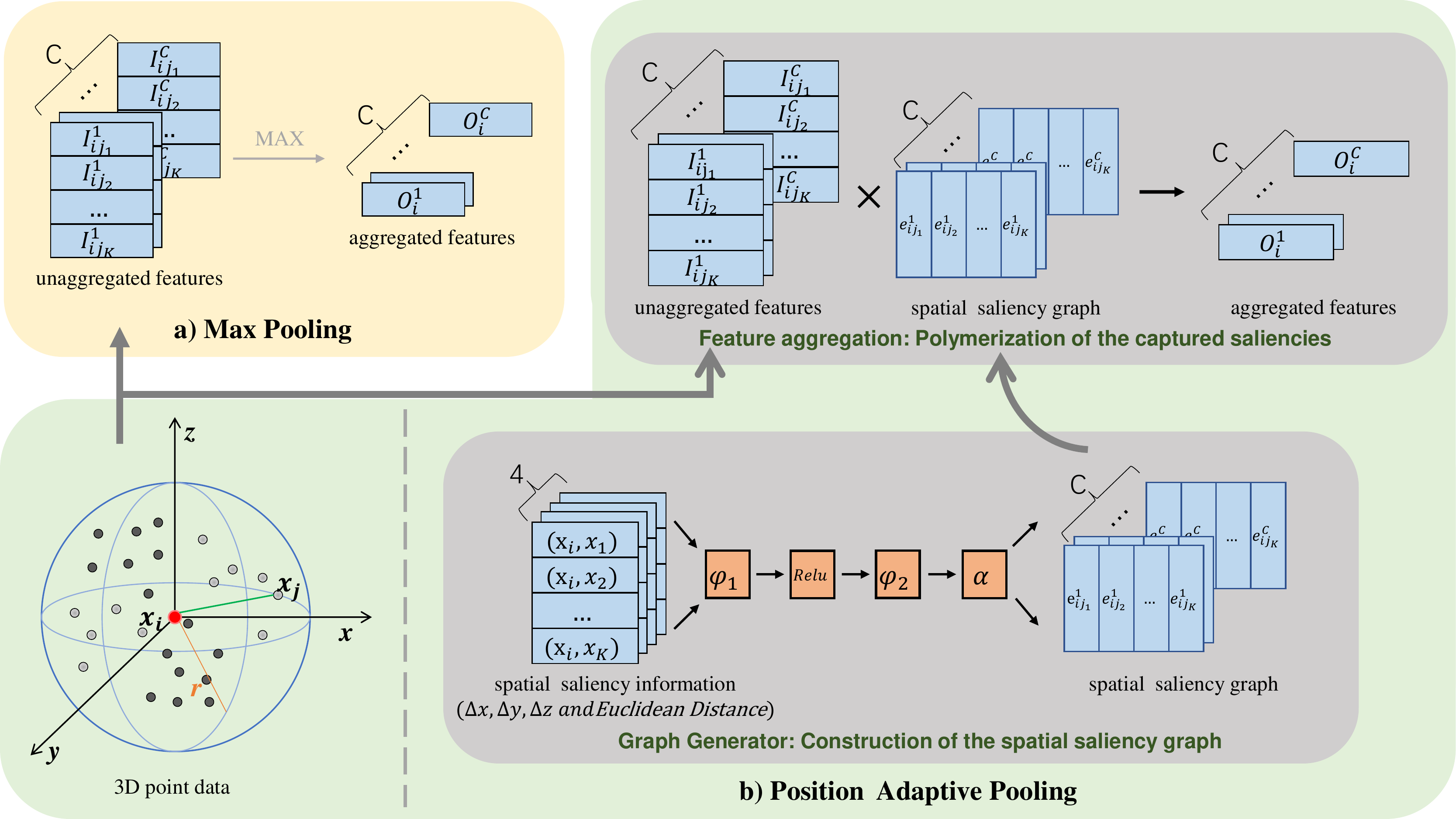} 
\caption{Overview of architecture of PAPooling. (a) shows the traditional feature aggregation operation in point clouds. (b) illustrates the workflow of our PAPooling, including two important modules, Graph Construction and Feature Aggregation. The first module, Graph Construction, is designed to take position features($dx,dy,dz$ and $Euclidean Distance$) to construct a spatial saliency graph of great fine-grained local geometric information, with channel-wise weights in its edges. The second module, Feature Aggregation module, is proposed to adjust features according to the channel-wise weights in the edges of graph and then converge the adjusted features in a method similar to Graph Convolution Network(GCN)}
\label{overview of PAPooling}
\end{figure*}

\section{Methods}
In this section, we first revisit common feature aggregation operation in point clouds. Then we introduce our Position Adaptive Pooling (PAPooling) and illustrate how to apply PAPooling to existing popular backbones as a plug-and-play end-to-end learnable layer and a originally designed pipeline. Finally, we compare PAPooling with prior relevant works.

\subsection{Review of conventional  aggregation operation}
In point clouds, dominant backbones such as PointNet++~\cite{qi2017pointnet++} and DGCNN~\cite{wang2019dynamicDGCNN} involve three essential strides: first splitting a raw point cloud into some local regions by a sampling method such as farthest point sampling (FPS) algorithm and a method of selecting neighboring points such as K-nearest neighbours (KNN) query or ball query, in the purpose of characterizing the original point cloud space by these key points, then extracting the corresponding feature of each selected local region by multi-layer perceptrons (MLPs)~\cite{hornik1991approximationMLP}, and finally aggregating all separate local region features into a global representation by means of simple max/average pooling.

Farthest point sampling (FPS) is widely used in many point cloud frameworks, as it can generate a relatively uniform sampled points. Therefore, their neighbors can cover all input point clouds as much as possible. Given $\mathcal{N}$ points in a point cloud $\mathcal{P} = \{ p_{i}\mid i=1,\dots,N \} \in{\mathbb{R}^{N \times 3 }}$, the output of Sampling and Grouping module is $\mathcal{F} = \{f_i\mid i=1,\dots,N_p\} \in{\mathbb{R}^{N_p \times K \times 3}}$, where $N_p$ and $K$ are artificial setting numbers respectively representing the number of key points in a point cloud and the number of neighboring points surrounding each key point, and $\mathbf{3}$ meaning the 3-dim coordinates.  Followling a multi-layer perceptrons (MLPs)~\cite{hornik1991approximationMLP} layer, point-to-point relations in the local region can be captured and the output feature map of $\mathcal{F}$ can be denote as $\mathcal{G} = \{g_i\mid i=1,\dots,N_p\} \in{\mathbb{R}^{N_p \times K \times C}}$, where ${C}$ is the number of output feature channels after MLPs. Evently these individual local features can be well concatenated via a feature aggregation operator.

Max pooling is a well-established operation with a wide range of usage in various areas. More significantly, it is the most commonly-used aggregation operator for point sets due to its superiority in addressing the  irregularity and disorderliness of point clouds. 
In a local region $\mathcal{G} = \{g_i\mid i=1,\dots,N_p\} \in{\mathbb{R}^{N_p \times K \times C}}$ of point cloud, there are many neighboring points $\mathcal{P} = \{p_j\mid j=1,\dots,K\} \in {\mathbb{R}^{K \times C}} $ surrounding ${i}-th$ local region $g_i$. $N_p$ means the number of local regions in point clouds and $K$ means the number of surrounding points in each local region. For $i-th$ local region $g_i$, the generalized max pooling can be formulated as:
\begin{equation}
    O(g_i)  = \max\limits_{j=1,2,\dots,K} (p_j)
\end{equation}
where $o_i=O(g_i) \in{\mathbb{R}^{1 \times C}}$ denotes $i-th$ local region feature after max pooling and $\mathcal{O} = \{o_i\mid i=1,\dots,N_p\} \in{\mathbb{R}^{N_p \times C}}$  denotes global region features aggregated by max pooling.


Just as shown in Fig.\ref{overview of PAPooling} a), there are $K$ unaggregated features in a feature channel of local region $I_i$. Through a feature aggregation like max pooling, only the most prominent feature can be retained, just like the $O_i$ shown in the figure. 


Max pooling enlarges the receptive field while inevitably introducing information loss such as follows: i) Feature's position information is completely lost after this step even though the convolution operation keeps them, because max pooling only retains the unique maximum value and ignores the position information;  ii) When some strong features appear repeatedly, max pooling can only keep a maximum value, and will lose other sub-strong features.

\subsection{Graph-based position-aware feature aggregation}
Max pooling only recklessly retains the most distinctive features in a channel while ignoring the other relatively weak features in the channel and the intrinsic connections between them, impairing the contextual relation-aware representation learning.
To address the shortcomings, we design Position Adaptive Pooling (PAPooling) alternative to max pooling, a brand new feature aggregation operator that leverages spatial distribution of point cloud via an innovative graph representation. As shown in Fig.\ref{overview of PAPooling} b), PAPooling comprises two essential modules, \textit{Graph Construction} and \textit{Feature Aggregation}, separately in charge of constructing a spatial saliency graph and aggregating features according to the graph, 
allowing the entire network well aggregate features adaptively, thus achieving fine-grained structure assembling.

\subsubsection{Graph Construction}
Considering the points in each local region are connected weakly in the spatial domain and graph is an excellent structure for representing the interconnections between nodes, we transfer all the points in a local region and their connections into a form of graph, where points corresponds to nodes and interactions to edges. 


The overview of Graph Construction is shown in Fig. \ref{overview of PAPooling} b). The key step of constructing the graph is to generate apposite edges between nodes by modeling interactions between points. We consider the spatial distribution extremely important resource, with superior performance in characterizing relationship between points. Spatial information will be represented in the form of relative positions as follows: 
\begin{equation}
    (x_i, x_{i}^j) = \mathop{\arg}\limits_{j = 1, 2, \dots, K}{ ((x_{i} - x_{i}^{j})   \oplus \mid\mid x_{i} - x_{i}^{j}  \mid\mid)}
\end{equation}


where $x_i$ and $x_i^j(j = 1, 2, \dots, K) $ respectively represents coordinates of center point and its K neighboring points in $i-th$ local region, $\oplus$ denotes the concatenation operator and $\mid\mid \cdot \mid\mid$ means calculating the Euclidean distance.

To construct a spatial saliency graph with generalized high-dimensional spatial information, we leverage relative positions with powerful attention mechanism to automatically learn important local features. Particularly in $i-th$ region, we customize a two-layer perception network to encode the relative positions for the purpose of augmenting the neighbouring point features, which is defined as: 
\begin{equation}
e_i = \alpha( \phi_2(ReLU( \phi_1(x_i, x_i^j) ))).
\end{equation}

${E} = \{ e_i\mid i = 1, \dots, N\}\in\mathbb{R}^{N\times K \times C} $ is the weight matrix in spatial saliency graph consisting of channel-wise attentive coefficients acting on unaggregated features. $\phi_m ( m=1, 2 ) $ indicates the parameters of multi-layer perceptrons(MLPs)~\cite{hornik1991approximationMLP} and $\alpha$ refers to  Softmax normalization operator. Softmax ensures the values derived from Equation $(3)$ lie in range $(0,1)$ and guarantees that each feature will be modified with a probability, with higher scores implying stronger relations between the input features and the weight matrix. Numbers of encoders $\phi$ and different normalization functions effect the ability to learn local structures. (See discussions in Section $\uppercase\expandafter{\romannumeral4}-D$). 
Specifically, local region of point cloud is a special graph that only has edges between the center node and its neighboring nodes, without  connections between neighboring nodes inside. 

\subsubsection{Feature Aggregation}
For the specificity of graph, accordingly we propose a graph-aggregation structure similar to Graph Convolution Networks (GCN) and adopt the high-order polynomial to model multi-scale interactions between nodes formulated as:
\begin{equation}
X^{out} = \sigma(\sum_{m=1}^{M} w_{m} \hat{E}_{m} X^{in} ),
\end{equation}
$X^{in} \in{\mathbb{R}^{C \times K \times 1 }}$ and $X^{out} \in{\mathbb{R}^{C \times 1 \times 1 }}$ are the input and output features in a local region, corresponding to $I_i$ and $O_i$ in Feature Aggregation module in Fig.\ref{overview of PAPooling} (b). $C$ is channels features, and $K$ denotes the number of nodes neighboring the center point in each local region. $m$ and $M$ are the polynomial order and total number of orders, denoting numbers of center nodes with neighboring nodes. $w_m$ is a trainable weight for the order $m$, $\hat{E}_{m}\in \mathbb{R}^{C \times 1 \times K}$ is the normalized matrix of graph constructed by Graph Construction, denotes the learnable parameter matrix for $m$-th order,  $\sigma$ is the activation function. 
Specifically, $M$ is 1 in the graph we constructed. Due to the particularity of point clouds that there exists a center point in a local region, constructed graph is very special that only has edges between the center node and its neighboring nodes, without  connections between neighboring nodes inside.

Through constructing this saliency graph and further performing saliency polymerization, our method can be able to reassemable feature converging process in a data-driven manner. With this dynamic constructing graph strategy, model gains ability of better aggregating geometry information and enriching the representation power of point cloud.

\subsection{Implementation into point-based models}
Dynamic position adaptive feature aggregation is a plug-and-play layer. It is a simple yet efficient method and generic enough to be applied in various point based models by simply replacing conventional max pooling. 

There are many different point cloud networks, yet most of them can be considered as different variants of the classical point based networks. To assess the effectiveness of PAPooling and minimize the impact from complicated network architectures, we choose two classical network backbones PointNet++~\cite{qi2017pointnet++} and DGCNN~\cite{wang2019dynamicDGCNN} as our baseline network architectures.

As mentioned before, PAPooling is utilized to replace the max pooling of the feature aggregation operators in PointNet++ and DGCNN. In addition, we only made minor adjustments to the network structure. In PointNet++, we retain the $MLP$s~\cite{hornik1991approximationMLP} and $FeaturePropagation$ layer, but replace the max pooling with our PAPooling. In DGCNN, we retain the $Egdeconv$ operator, but replace the max pooling and average pooling with our PAPooing. In brief, we make little changes to the structure of backbone, only change the way of feature aggregation, replacing max pooling or average pooling with PAPooing. Except for the only layer change in network architecture, the other training strategies are exactly the same as baselineworks. The comparison is apple to apple.

\subsection{Comparison with Prior Works}
\subsubsection{Comparison with RandLA-Net~\cite{hu2020randla-net}}
RandLA-Net designs an Computing Attention Scores to learn a unique attention score and associate them with features to generate an informative feature vector. However, the operator uses simple weighted summation methods to exploit the weights, which is limited in realizing the potential of these attentive weights. Our PAPooling, nevertheless, utlizes Graph Convolution Network (GCN) to combine weights from graph and obtain better performance in learn local local geometric information.  

\subsubsection{Comparison with PosPool~\cite{liu2020closer_pospool}}
PosPool proposes a deep residual architecture~\cite{he2016deep_resnet} for point cloud and strives to design a simple pooling way. It simply multiplies the relative position and features, lacking further leverage of geometric information. Whereas, PAPooling extracts high-level geometric information and aggregates features in a novel graph representation, preserving fine-grained local geometry.

\begin{table}[ht]
\centering
\begin{tabular}{lcc}
\toprule
    Method & Input & Acc(\%)  \\
    \hline
    MVCNN~\cite{MVCNN}   & multi-view & 90.1 \\
    OctNet~\cite{riegler2017octnet} & hybird grid octree & 86.5\\
    PointwiseCNN~\cite{hua2018pointwise} & 1K points & 86.1\\
    PointNet~\cite{qi2017pointnet}  & 1K points & 89.2 \\
    PCNN~\cite{PCNN}  & 1K points & 92.3 \\
    PointCNN~\cite{PointCNN}  & 1K points & 92.5 \\
    PointWeb~\cite{PointWeb}  & 1K points+normal & 92.3 \\
    PointConv~\cite{PointConv}  & 1K points+normal & 92.5 \\
    RS-CNN w/o vot.~\cite{liu2019relation_rscnn} & 1K points & 92.4 \\
    RS-CNN w vot.~\cite{liu2019relation_rscnn} & 1K points & 93.6 \\    
    KPConv~\cite{thomas2019kpconv} & 1K points & 92.9 \\
    DensePoint~\cite{DensePoint}  & 1K points & 93.2 \\
    Point2Node~\cite{Point2Node} & 1K points & 93.0 \\
    FPConv~\cite{lin2020fpconv} & 1K points & 92.5 \\
    PosPool~\cite{liu2020closer_pospool} & 5k points & 93.2\\
    PAConv w/o vot.~\cite{xu2021paconv} & 1K points & 93.6\\
    PAConv w vot.~\cite{xu2021paconv} & 1K points & 93.9\\
    \hline
    PointNet++~\cite{qi2017pointnet++} & 1K points & 90.7 \\
    Ours(*PointNet++) w/o vot. & 1K points & 92.7 (\textbf{2.0} $\uparrow$) \\
    \hline
    DGCNN~\cite{wang2019dynamicDGCNN} & 1K points & 92.9 \\    
    Ours(*DGCNN) w/o vot. & 1K points & \textbf{93.2} (\textbf{0.3} $\uparrow$) \\
\bottomrule
\end{tabular}
\caption{Classification accuracy on ModelNet40. *PointNet++ and *DGCNN respectively denote using PointNet++ and DGCNN as the backbones. “vot.” indicates multi-scale inference. PAPooling obviously improves the performance of baseline and DGCNN embedded with PAPooling surpasses other methods. }
\vspace{-2pt}
\label{table1}
\end{table}

\section{Experiments}
To evaluate the behavior of PAPooling, we conduct comprehensive experiments on tasks of object classification, shape part segmentation and indoor scene segmentation. First we describe our baseline network architectures and corresponding implementations for these tasks, and then extend into networks embedded with our proposed PAPooling. Further, we conduct extensive ablation studies to verify the effectiveness of PAPooling in detail. 

\subsection{Object Classification}
\noindent\textbf{Dateset}
First we evaluate our model on ModelNet40~\cite{referpaper27modelnet40} for object classification. It consists of 3D meshed models from 40 categories, with 9843 training models and 2468 testing models.

\noindent\textbf{Implementation Details} 
Here we choose PointNet++~\cite{qi2017pointnet++} and DGCNN~\cite{wang2019dynamicDGCNN} as our baseline. In PointNet++, we set training batch size to 24 and epochs to 200. We use Adam optimizer, set learning rate to 0.001 and decay rate to 0.0001.
In DGCNN, we set training batch size to 32 and epochs to 250. We train the model by SGD optimizer with momentum of 0.9, set learning rate to 0.1 and decay rate to 0.0001. 

\begin{table}[t]
\centering
\begin{tabular}{lcc}
\toprule
    Method & Cls. mIoU(\%) & Ins. mIoU(\%)   \\
    \hline
    PointNet~\cite{qi2017pointnet} & 80.4 & 83.7 \\
    SynspecCNN~\cite{yi2017syncspeccnn} & 82.0 & 84.7 \\
    PCNN~\cite{PCNN} & 81.8 & 85.1 \\
    SpiderCNN~\cite{xu2018spidercnn} & 82.4 & 85.3 \\
    PointCNN~\cite{PointCNN} & 84.6 & 86.1 \\
    PointConv~\cite{PointConv} & 82.8 & 85.7 \\
    RS-CNN w/o vot.~\cite{liu2019relation_rscnn} & 84.2 & 85.8 \\
    RS-CNN w/ vot.~\cite{liu2019relation_rscnn} & 84.0 & 86.2 \\
    KPConv~\cite{thomas2019kpconv} & 85.1 & \textbf{86.4} \\
    PAConv~\cite{xu2021paconv} & 84.6 & 86.1 \\
    \hline
    PointNet++~\cite{qi2017pointnet++} & 81.9 & 85.1  \\
    Ours(*PointNet++) w/o vot. & 82.2(\textbf{0.3} $\uparrow$) & 85.3(\textbf{0.2}$\uparrow$)  \\
    \hline
    DGCNN~\cite{wang2019dynamicDGCNN} & 82.3 & 85.2 \\
    Ours(*DGCNN) w/o vot.  & 83.8(\textbf{1.5} $\uparrow$) & \textbf{85.6}(\textbf{0.4} $\uparrow$)  \\
\bottomrule
\end{tabular}
\caption{Part segmentation results on ShapeNet part dataset. *PointNet++ and *DGCNN respectively denote using PointNet++ and DGCNN as the backbones. “vot.” indicates multi-scale inference. Our method  improves the performance of PointNet++ and DGCNN, especially class mIoU on DGCNN. }
\label{table2}
\end{table}

\begin{figure}[htbp]
\centering
\includegraphics[width=1.0\columnwidth]{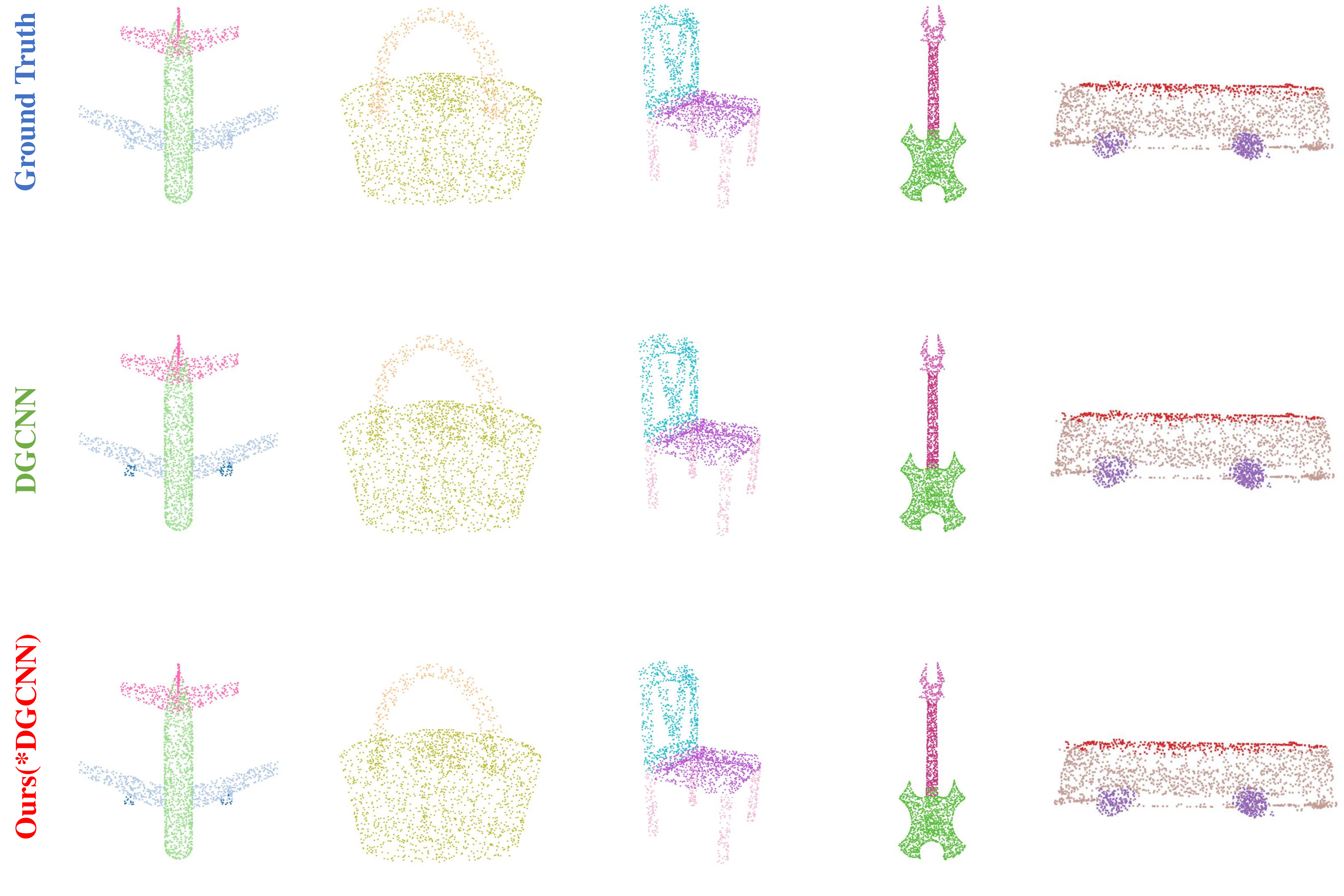} 
\caption{Visualization of shape part segmentation results on ShapeNet Parts. The first row is ground truth. The second and third row respectively shows the performance of baseline (DGCNN) and our PAPooling}
\label{fig3}
\end{figure}

\noindent\textbf{Result}
Table.\ref{table1} shows the results for the classification task. PAPooling significantly improves the classification accuracy with 2.0\% on PointNet++ and 0.3\% on DGCNN. Especially, the accuracy achieved by DGCNN+PAPooling is 93.2\%, which outperforms many backbones. Unlike other backbones that use voting to get higher performance, we didn’t take voting and still get good result. The results without voting better reflects the performance gained purely from model designs and show the effectiveness of our PAPooling.

\noindent\textbf{Speed}
Notably, the memory footprint of PointNet++ is 3567MB and that of PointNet++ embedded with PAPooling is 5900MB, and the inference time for one picture in PointNet++ is 0.54s and that of PointNet++ embedded PAPooling is 0.57s, which demonstrates that PAPooling is efficient, with very minimal computational overhead.

\subsection{Shape Part Segmentation}
\noindent\textbf{Dateset} 
Our method is also evaluated on ShapeNet Part~\cite{referpaper28_shapenet} for shape part segmentation. It contains 16881 shapes with 16 categories and is labeled in 50 parts where each shape has 2-5 parts. 2048 points are sampled from each shape and each point is annotated with a part label.

\noindent\textbf{Implementation Details} 
We choose PointNet++~\cite{qi2017pointnet++} and DGCNN~\cite{wang2019dynamicDGCNN} as our baseline. In PointNet++, we set training batch size to 16 and epochs to 250. We use Adam optimizer, set learning rate to 0.001 and decay rate to 0.0001. In DGCNN, we set training batch size to 32 and epochs to 200. We train the model by SGD optimizer with momentum of 0.9, set learning rate to 0.1 and decay rate to 0.0001.

\noindent\textbf{Result}
Table.\ref{table2} lists the instance average and class average mean Inter-over-Union (mIoU). PAPooling lifts performance of PointNet++ on class mIoU ($\uparrow0.3\%$) and instance mIoU ($\uparrow0.2\%$). Notably,  PAPooling lifts the performance of DGCNN on both class mIoU ($\uparrow2.3\%$) and instance mIoU ($\uparrow0.9\%$). Embedded with PAPooling, DGCNN performs 83.5\% in Cls. mIoU and 85.6\% in Ins. mIoU, better than many well-known models,  with much lower computational efficiency. 

Fig.\ref{fig3} visualizes ground truth and segmentation results of our method. As we can see from figure, our method has outstanding performance in shape part segmentation task.

\subsection{Indoor Scene Segmentation}

\noindent\textbf{Dataset} 
Large-scale indoor scene segmentation is a challenging task. To further assess our method, we employ our PAPooling in Stanford 3D Indoor Space (S3DIS)~\cite{referpaper29_s3dis} dataset, which includes 271 rooms in 6 areas. 273 million points are scanned from 3 different buildings, and each point is annotated with one semantic label from 13 classes.

\noindent\textbf{Implementation Details} 
We choose PointNet++~\cite{qi2017pointnet++} and DGCNN~\cite{wang2019dynamicDGCNN} as our baseline. In PointNet++, we set training batch size to 16 and epochs to 32. We use Adam optimizer, set learning rate to 0.001 and decay rate to 0.0001. In DGCNN, we set training batch size to 32 and epochs to 100. We train the model by SGD optimizer with momentum of 0.9, set learning rate to 0.1 and decay rate to 0.0001.

\noindent\textbf{Result}
For the evaluation table, we use mean of classwise mean Inter-over-Union (mIoU). As shown in Tab.\ref{table3}, PAPooling considerably promotes PointNet++ by 1\% and DGCNN by 4.2\%. DGCNN embedded with our PAPooling achieves a very robust mIoU of 60.3\%, which is an excellent example of advancement of PAPooling in precisely segmenting objects.

\begin{figure*}[h]
\centering
\includegraphics[width=1.0\textwidth]{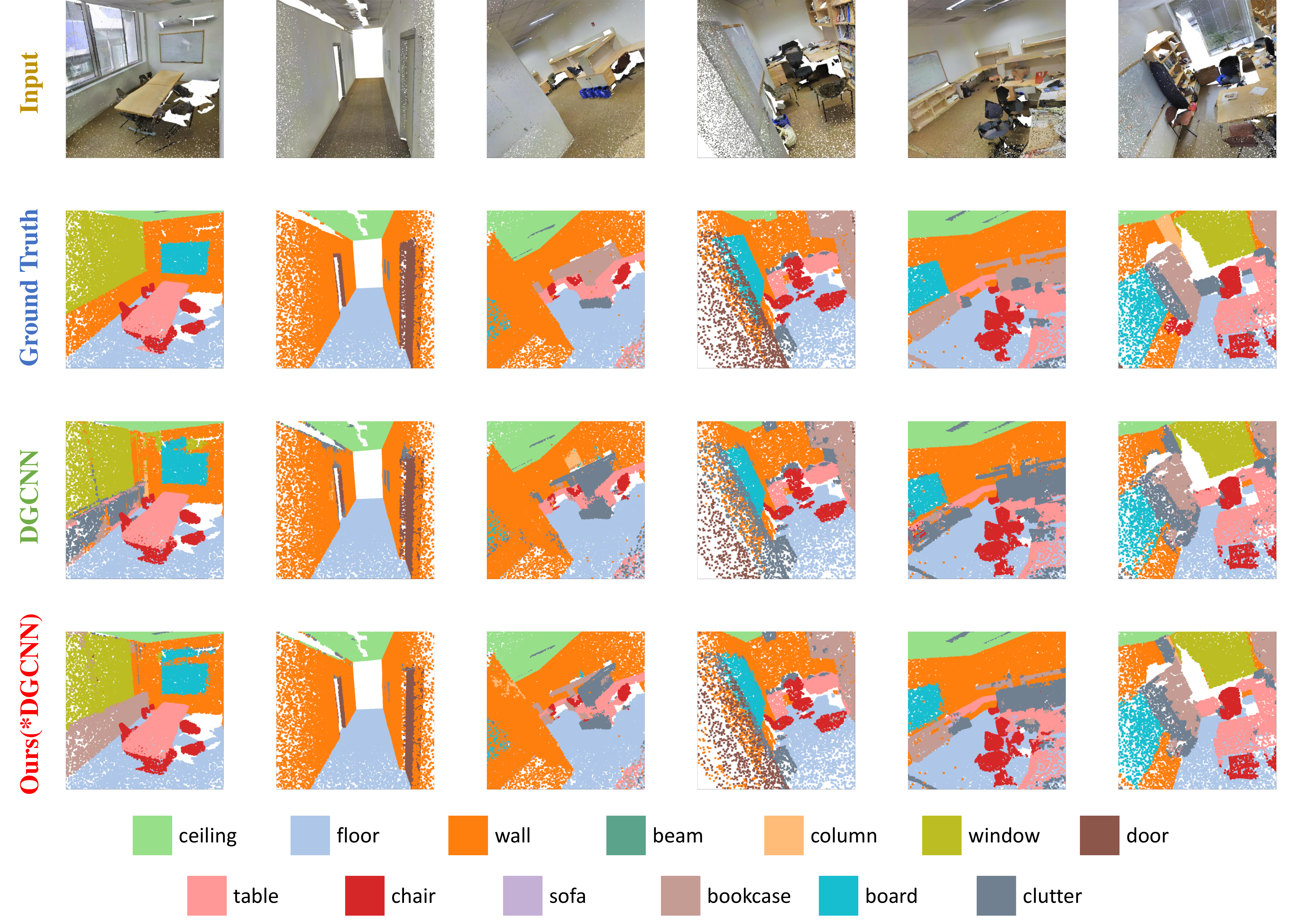} 
\caption{Visualization of semantic segmentation results on S3DIS Area-5. The first row shows original scene inputs, the second row shows the ground truth annotations, and the third and the last row respectively shows the scenes segmented by DGCNN and our method. Each column denotes a scene in S3DIS Area-5.}
\label{fig4}
\end{figure*}

The visualization of indoor scene segmentation results are shown in Fig.\ref{fig4}. As we can see from the figure, embedded with our method, DGCNN performs well on S3DIS dataset. At the boundary of objects, our method is obviously superior to baseline, effectively verifies that our method can capture fine-grained geometric information.   

\begin{table}[h]
\centering
\begin{tabular}{lc}
\toprule
    Method & mIoU(\%)   \\
    \hline
    PointNet~\cite{qi2017pointnet} & 41.1 \\
    SegCloud~\cite{tchapmi2017segcloud} & 48.9 \\
    TangentConv~\cite{tatarchenko2018tangent_tangentConv} & 52.6 \\
    PointCNN~\cite{PointCNN} & 57.3\\
    ParamConv~\cite{wang2018deep_paramconv} & 58.3 \\
    SPG~\cite{landrieu2018large_spg} & 58.0 \\
    PointWeb~\cite{PointWeb} & 60.2 \\
    PCNN~\cite{PCNN} & 58.2 \\
    FPConv~\cite{lin2020fpconv} & 62.8 \\
    Point2Node~\cite{Point2Node} & 62.9 \\
    KPConv rigid~\cite{thomas2019kpconv} & 65.4 \\
    KPConv deform~\cite{thomas2019kpconv} & \textbf{67.1} \\
    PosPool~\cite{liu2020closer_pospool} & 66.7 \\
    \hline
    PointNet++~\cite{qi2017pointnet++} & 53.5 \\
    Ours(*PointNet++) & 54.5(\textbf{1.0} $\uparrow$) \\
    \hline
    DGCNN~\cite{wang2019dynamicDGCNN} & 56.1 \\
    Ours(*DGCNN) & 60.3(\textbf{4.2} $\uparrow$)  \\
\bottomrule
\end{tabular}
\caption{3D semantic segmentation results on S3DIS dataset, *PointNet++ and *DGCNN respectively denote using PointNet++ and DGCNN as the backbones.  }
\label{table3}
\end{table}

\subsection{Ablation Studies} 
To better comprehend PAPooling, we conduct ablation studies on Objection Classification on ModelNet40 dataset.

\subsubsection{Weights of different neighboring points}
\begin{figure}[h]
\centering
\includegraphics[width=1.0\columnwidth]{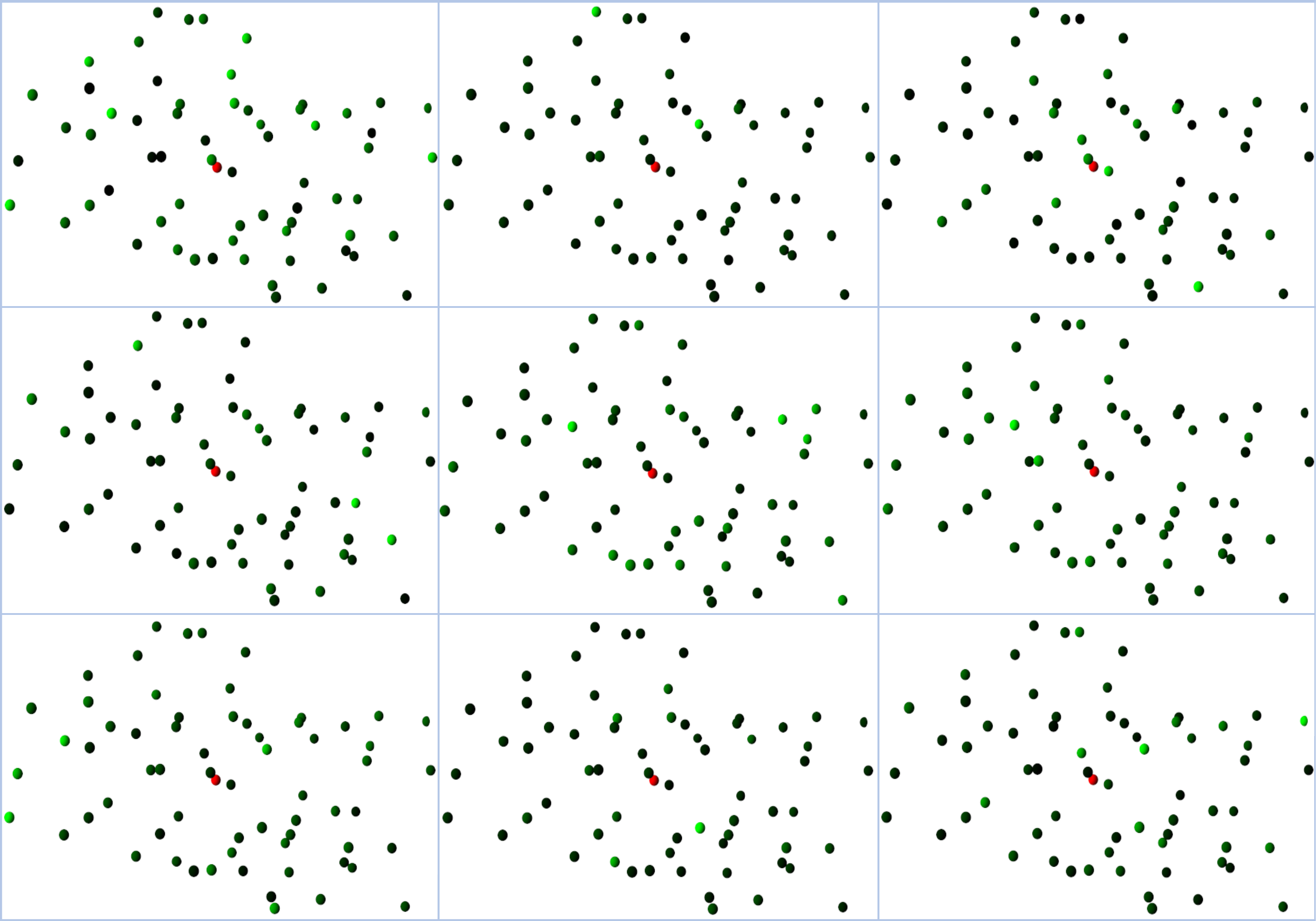} 
\caption{Visualization of weights of 64 neighboring points in a local region. Here we choose 9 different channels to compare the contribution of each point. The weights of nodes are related to the colors, nodes of brighter colors have larger weights. }
\label{fig_weights.}
\end{figure}

To better understand the inner mechanism of Graph Construction and explore how much contribution each point makes in a local region, we decide to visualize the weights in edges of graph. For convenience, we omit the edges of graph, but leave only nodes. We draw these nodes with green of different brightness, the brightness of nodes is related to the weight, bright nodes have high value of weights and dark nodes have low value of weights. Here we visualize a local region with 64 neighboring points from 9 different channels to compare the contribution of each point. Fig.\ref{fig_weights.} shows that for each point of each feature, there is a corresponding distinct weight to adjust it. By this way, all the features get modified dynamically. Then PAPooling can easily aggregate feature by converging these  dynamically improved features.

\subsubsection{Weights normalization}
We also investigate widely-used normalization functions in order to adjust the score distribution. Table.\ref{table4} shows that Softmax normalization outperforms other schemes. It suggests that predicting scores for all weight matrices as a whole (Softmax) is superior than considering each score independently (Sigmoid and Tanh). 
Besides, to test if shared weights in constructed graph will impair the performance of algorithm, we try to make Graph Construction generate channel-wise and channel-agnostic weights(by setting the output channels of last MLP to one) respectively. The results indicate that channel-wise weights have advantages in paying attention to information exchange between channels.

\begin{table}[t]
\centering
\begin{tabular}{lcc}
\toprule
    Normalization Function & Accuracy  \\
    \hline
    w/o normalization  & 91.93 \\
    Sigmoid & 91.86 \\
    Tanh & 92.13 \\
    Logsoftmax & 92.36 \\
     \hline
      Softmax (Channel-agnostic) & 92.20 \\
     \textbf{Softmax (Channel-wise}) & \textbf{92.74} \\
\bottomrule
\end{tabular}
\caption{Classification results on ModelNet40 using PAPooling with different normalization functions. Normalization functions control the score distribution and results of feature aggregation according to constructed graph.}
\label{table4}
\end{table}

\begin{figure}[!h]
\centering
\includegraphics[width=1.0\columnwidth]{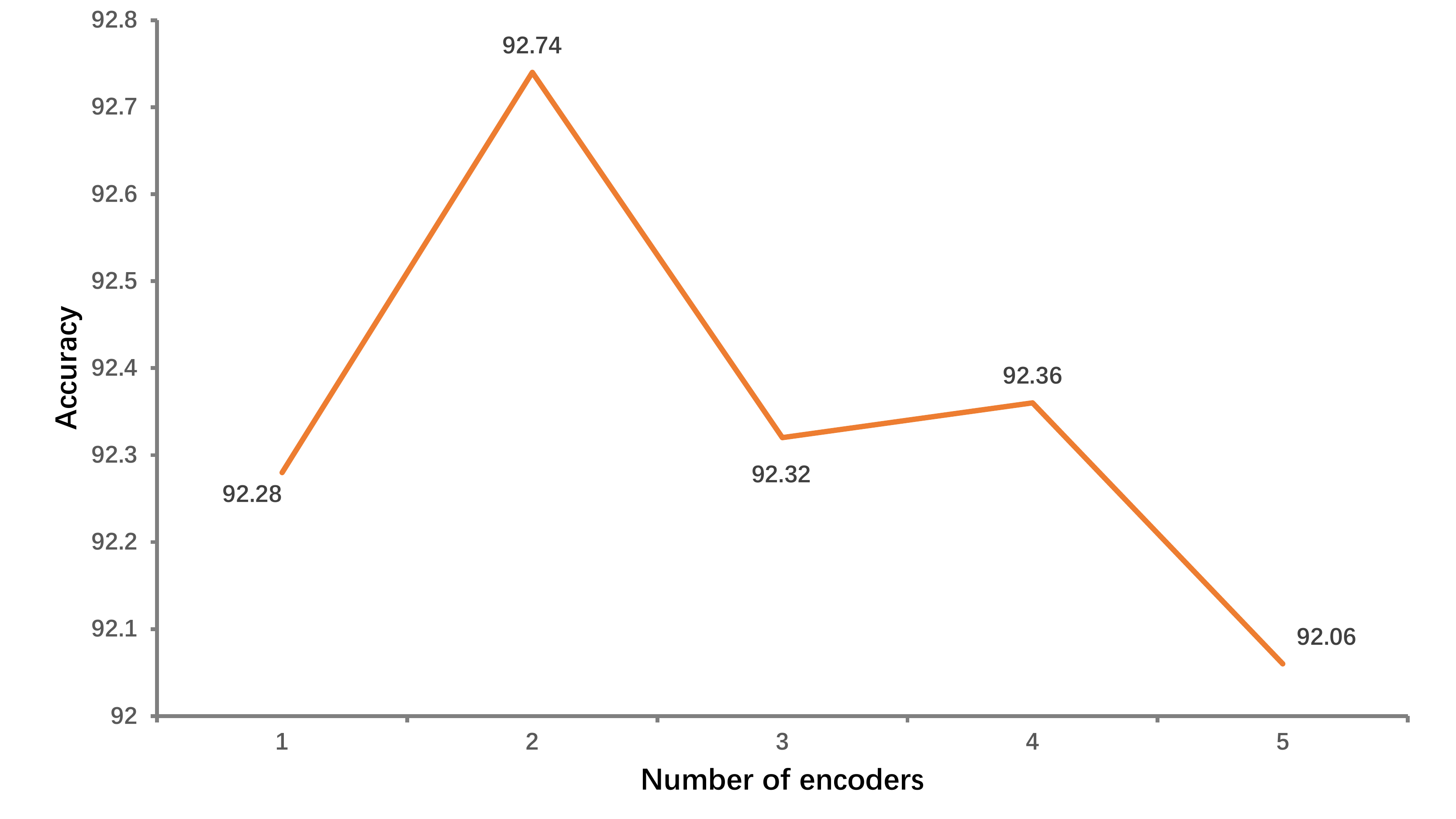} 
\caption{Visualization of the change of accuracy as the number of encoders in Graph Construction increases.}
\vspace{+4mm}
\label{numberofconv2d}
\end{figure}

\subsubsection{The number of encoders in Graph Construction}
We further conduct experiments to figure out the influence of the number of encoders in Graph Construction. As shown in Fig.\ref{numberofconv2d}, when the number of encoders increases, according results don't increase consistently and model achieves the highest accuracy when the number is 2, which indicates that simple design of Graph Construction can achieve ideal results. This definitely demonstrates the power of our proposed approach. Finally, we achieve the best and most stable performance when the number is 2.

\section{Conclusion}
We present PAPooling, a feature aggregation operator with dynamic learning spatial distribution of point cloud data. Through constructing spatial saliency graph and further performing saliency polymerization, our method reassemables feature converging process in a data-driven manner. With this dynamic constructing graph strategy, PAPooling gains the ability of better aggregating geometry information and enriching the representation power of point cloud. Extensive experiments and ablation studies illustrate the effectiveness of our PAPooling.

\bibliographystyle{IEEEtran}
\bibliography{reference.bib}

\vfill



\end{document}